\documentclass{article}

\usepackage{PRIMEarxiv}

\usepackage{threeparttable} 
\usepackage[utf8]{inputenc} % allow utf-8 input
\usepackage[T1]{fontenc}    % use 8-bit T1 fonts
\usepackage{hyperref}       % hyperlinks
\usepackage{url}            % simple URL typesetting
\usepackage{booktabs}       % professional-quality tables
\usepackage{amsfonts}       % blackboard math symbols
\usepackage{nicefrac}       % compact symbols for 1/2, etc.
\usepackage{microtype}      % microtypography
\usepackage{lipsum}
\usepackage{fancyhdr}       % header
\usepackage{graphicx}       % graphics
\graphicspath{{media/}}     % organize your images and other figures under media/ folder
\usepackage{amsmath}
\title{Zep: A Temporal Knowledge Graph Architecture for Agent Memory
}

\author{
  Preston Rasmussen \\
  Zep AI \\
  \texttt{preston@getzep.com} \\
   \And
  Pavlo Paliychuk  \\
  Zep AI \\
  \texttt{paul@getzep.com} \\
     \And
  Travis Beauvais \\
  Zep AI \\
  \texttt{travis@getzep.com} \\   \And
  Jack Ryan  \\
  Zep AI \\
  \texttt{jack@getzep.com} \\   \And
  Daniel Chalef  \\
  Zep AI \\
  \texttt{daniel@getzep.com} \\
}

\begin{document}
\maketitle

\begin{abstract}
We introduce Zep, a novel memory layer service for AI agents that outperforms the current state-of-the-art system, MemGPT, in the Deep Memory Retrieval (DMR) benchmark. Additionally, Zep excels in more comprehensive and challenging evaluations than DMR that better reflect real-world enterprise use cases. While existing retrieval-augmented generation (RAG) frameworks for large language model (LLM)-based agents are limited to static document retrieval, enterprise applications demand dynamic knowledge integration from diverse sources including ongoing conversations and business data. Zep addresses this fundamental limitation through its core component Graphiti---a temporally-aware knowledge graph engine that dynamically synthesizes both unstructured conversational data and structured business data while maintaining historical relationships. In the DMR benchmark, which the MemGPT team established as their primary evaluation metric, Zep demonstrates superior performance (94.8\% vs 93.4\%). Beyond DMR, Zep's capabilities are further validated through the more challenging LongMemEval benchmark, which better reflects enterprise use cases through complex temporal reasoning tasks. In this evaluation, Zep achieves substantial results with accuracy improvements of up to 18.5\% while simultaneously reducing response latency by 90\% compared to baseline implementations. These results are particularly pronounced in enterprise-critical tasks such as cross-session information synthesis and long-term context maintenance, demonstrating Zep's effectiveness for deployment in real-world applications.
\end{abstract}

% keywords can be removed
% \keywords{First keyword \and Second keyword \and More}

\section{Introduction}
The impact of transformer-based large language models (LLMs) on industry and research communities has garnered significant attention in recent years \cite{attentionneed}. A major application of LLMs has been the development of chat-based agents. However, these agents' capabilities are limited by the LLMs' context windows, effective context utilization, and knowledge gained during pre-training. Consequently, additional context is required to provide out-of-domain (OOD) knowledge and reduce hallucinations.

Retrieval-Augmented Generation (RAG) has emerged as a key area of interest in LLM-based applications. RAG leverages Information Retrieval (IR) techniques pioneered over the last fifty years\cite{sparckjones1972} to supply necessary domain knowledge to LLMs.

Current approaches using RAG have focused on broad domain knowledge and largely static corpora—that is, document contents added to a corpus seldom change. For agents to become pervasive in our daily lives, autonomously solving problems from trivial to highly complex, they will need access to a large corpus of continuously evolving data from users' interactions with the agent, along with related business and world data. We view empowering agents with this broad and dynamic "memory" as a crucial building block to actualize this vision, and we argue that current RAG approaches are unsuitable for this future. Since entire conversation histories, business datasets, and other domain-specific content cannot fit effectively inside LLM context windows, new approaches need to be developed for agent memory. Adding memory to LLM-powered agents isn't a new idea—this concept has been explored previously in MemGPT \cite{memgpt}.

Recently, Knowledge Graphs (KGs) have been employed to enhance RAG architectures to address many of the shortcomings of traditional IR techniques\cite{graphrag}. In this paper, we introduce Zep\cite{zep}, a memory layer service powered by Graphiti\cite{graphiti}, a dynamic, temporally-aware knowledge graph engine. Zep ingests and synthesizes both unstructured message data and structured business data. The Graphiti KG engine dynamically updates the knowledge graph with new information in a non-lossy manner, maintaining a timeline of facts and relationships, including their periods of validity. This approach enables the knowledge graph to represent a complex, evolving world.

As Zep is a production system, we've focused heavily on the accuracy, latency, and scalability of its memory retrieval mechanisms. We evaluate these mechanisms' efficacy using two existing benchmarks: a Deep Memory Retrieval task (DMR) from MemGPT\cite{memgpt}, as well as the LongMemEval benchmark\cite{longmemeval}.

\section{Knowledge Graph Construction}
In Zep, memory is powered by a temporally-aware dynamic knowledge graph $\mathcal{G}=(\mathcal{N}, \mathcal{E}, \phi)$, where $\mathcal{N}$ represents nodes, $\mathcal{E}$ represents edges, and $\phi:\mathcal{E}\to\mathcal{N}\times\mathcal{N}$ represents a formal incidence function. This graph comprises three hierarchical tiers of subgraphs: an episode subgraph, a semantic entity subgraph, and a community subgraph.
\begin{itemize}
\item \textbf{Episode Subgraph} $\mathcal{G}_e$: Episodic nodes (episodes), $n_i\in\mathcal{N}_e$, contain raw input data in the form of messages, text, or JSON. Episodes serve as a non-lossy data store from which semantic entities and relations are extracted. Episodic edges, $e_i\in\mathcal{E}_e\subseteq\phi^*(\mathcal{N}_e\times\mathcal{N}_s)$, connect episodes to their referenced semantic entities.
\item \textbf{Semantic Entity Subgraph} $\mathcal{G}_s$: The semantic entity subgraph builds upon the episode subgraph. Entity nodes (entities), $n_i\in\mathcal{N}_s$, represent entities extracted from episodes and resolved with existing graph entities. Entity edges (semantic edges), $e_i\in\mathcal{E}_s\subseteq\phi^*(\mathcal{N}_s\times\mathcal{N}_s)$, represent relationships between entities extracted from episodes.
\item \textbf{Community Subgraph} $\mathcal{G}_c$: The community subgraph forms the highest level of Zep's knowledge graph. Community nodes (communities), $n_i\in\mathcal{N}_c$, represent clusters of strongly connected entities. Communities contain high-level summarizations of these clusters and represent a more comprehensive, interconnected view of $\mathcal{G}_s$'s structure. Community edges, $e_i\in\mathcal{E}_c\subseteq\phi^*(\mathcal{N}_c\times\mathcal{N}_s)$, connect communities to their entity members.
\end{itemize}

The dual storage of both raw episodic data and derived semantic entity information mirrors psychological models of human memory. These models distinguish between episodic memory, which represents distinct events, and semantic memory, which captures associations between concepts and their meanings \cite{gonzalez}. This approach enables LLM agents using Zep to develop more sophisticated and nuanced memory structures that better align with our understanding of human memory systems. Knowledge graphs provide an effective medium for representing these memory structures, and our implementation of distinct episodic and semantic subgraphs draws from similar approaches in AriGraph \cite{arigraph}.

Our use of community nodes to represent high-level structures and domain concepts builds upon work from GraphRAG \cite{graphrag}, enabling a more comprehensive global understanding of the domain. The resulting hierarchical organization—from episodes to facts to entities to communities—extends existing hierarchical RAG strategies \cite{chen2024hiqahierarchicalcontextualaugmentation}\cite{goel2024hirohierarchicalinformationretrieval}.

\subsection{Episodes}
Zep's graph construction begins with the ingestion of raw data units called Episodes. Episodes can be one of three core types: message, text, or JSON. While each type requires specific handling during graph construction, this paper focuses on the message type, as our experiments center on conversation memory. In our context, a message consists of relatively short text (several messages can fit within an LLM context window) along with the associated actor who produced the utterance.

Each message includes a reference timestamp $t_\text{ref}$ indicating when the message was sent. This temporal information enables Zep to accurately identify and extract relative or partial dates mentioned in the message content (e.g., "next Thursday," "in two weeks," or "last summer"). Zep implements a bi-temporal model, where timeline $T$ represents the chronological ordering of events, and timeline $T'$ represents the transactional order of Zep's data ingestion. While the $T'$ timeline serves the traditional purpose of database auditing, the $T$ timeline provides an additional dimension for modeling the dynamic nature of conversational data and memory. This bi-temporal approach represents a novel advancement in LLM-based knowledge graph construction and underlies much of Zep's unique capabilities compared to previous graph-based RAG proposals.

The episodic edges, $\mathcal{E}_e$, connect episodes to their extracted entity nodes. Episodes and their derived semantic edges maintain bidirectional indices that track the relationships between edges and their source episodes. This design reinforces the non-lossy nature of Graphiti's episodic subgraph by enabling both forward and backward traversal: semantic artifacts can be traced to their sources for citation or quotation, while episodes can quickly retrieve their relevant entities and facts. While these connections are not directly examined in this paper's experiments, they will be explored in future work.

\subsection{Semantic Entities and Facts}
\subsubsection{Entities}
ntity extraction represents the initial phase of episode processing. During ingestion, the system processes both the current message content and the last $n$ messages to provide context for named entity recognition. For this paper and in Zep's general implementation, $n=4$, providing two complete conversation turns for context evaluation. Given our focus on message processing, the speaker is automatically extracted as an entity. Following initial entity extraction, we employ a reflection technique inspired by reflexion\cite{reflexion} to minimize hallucinations and enhance extraction coverage. The system also extracts an entity summary from the episode to facilitate subsequent entity resolution and retrieval operations.

After extraction, the system embeds each entity name into a 1024-dimensional vector space. This embedding enables the retrieval of similar nodes through cosine similarity search across existing graph entity nodes. The system also performs a separate full-text search on existing entity names and summaries to identify additional candidate nodes. These candidate nodes, together with the episode context, are then processed through an LLM using our entity resolution prompt. When the system identifies a duplicate entity, it generates an updated name and summary.

Following entity extraction and resolution, the system incorporates the data into the knowledge graph using predefined Cypher queries. We chose this approach over LLM-generated database queries to ensure consistent schema formats and reduce the potential for hallucinations.

Selected prompts for graph construction are provided in the appendix.

\subsubsection{Facts}
or each fact containing its key predicate. Importantly, the same fact can be extracted multiple times between different entities, enabling Graphiti to model complex multi-entity facts through an implementation of hyper-edges.

Following extraction, the system generates embeddings for facts in preparation for graph integration. The system performs edge deduplication through a process similar to entity resolution. The hybrid search for relevant edges is constrained to edges existing between the same entity pairs as the proposed new edge. This constraint not only prevents erroneous combinations of similar edges between different entities but also significantly reduces the computational complexity of the deduplication process by limiting the search space to a subset of edges relevant to the specific entity pair.

\subsubsection{Temporal Extraction and Edge Invalidation}
A key differentiating feature of Graphiti compared to other knowledge graph engines is its capacity to manage dynamic information updates through temporal extraction and edge invalidation processes.

The system extracts temporal information about facts from the episode context using $t_\text{ref}$. This enables accurate extraction and datetime representation of both absolute timestamps (e.g., "Alan Turing was born on June 23, 1912") and relative timestamps (e.g., "I started my new job two weeks ago"). Consistent with our bi-temporal modeling approach, the system tracks four timestamps: $t'\text{created}$ and $t'\text{expired} \in T'$ monitor when facts are created or invalidated in the system, while $t_\text{valid}$ and $t_\text{invalid} \in T$ track the temporal range during which facts held true. These temporal data points are stored on edges alongside other fact information.

The introduction of new edges can invalidate existing edges in the database. The system employs an LLM to compare new edges against semantically related existing edges to identify potential contradictions. When the system identifies temporally overlapping contradictions, it invalidates the affected edges by setting their $t_\text{invalid}$ to the $t_\text{valid}$ of the invalidating edge. Following the transactional timeline $T'$, Graphiti consistently prioritizes new information when determining edge invalidation.

This comprehensive approach enables the dynamic addition of data to Graphiti as conversations evolve, while maintaining both current relationship states and historical records of relationship evolution over time.

\subsection{Communities}
After establishing the episodic and semantic subgraphs, the system constructs the community subgraph through community detection. While our community detection approach builds upon the technique described in GraphRAG\cite{graphrag}, we employ a label propagation algorithm \cite{labelprop} rather than the Leiden algorithm \cite{leiden}. This choice was influenced by label propagation's straightforward dynamic extension, which enables the system to maintain accurate community representations for longer periods as new data enters the graph, delaying the need for complete community refreshes.

The dynamic extension implements the logic of a single recursive step in label propagation. When the system adds a new entity node $n_i\in\mathcal{N}_s$ to the graph, it surveys the communities of neighboring nodes. The system assigns the new node to the community held by the plurality of its neighbors, then updates the community summary and graph accordingly. While this dynamic updating enables efficient community extension as data flows into the system, the resulting communities gradually diverge from those that would be generated by a complete label propagation run. Therefore, periodic community refreshes remain necessary. However, this dynamic updating strategy provides a practical heuristic that significantly reduces latency and LLM inference costs.

Following \cite{graphrag}, our community nodes contain summaries derived through an iterative map-reduce-style summarization of member nodes. However, our retrieval methods differ substantially from GraphRAG's map-reduce approach \cite{graphrag}. To support our retrieval methodology, we generate community names containing key terms and relevant subjects from the community summaries. These names are embedded and stored to enable cosine similarity searches.

\section{Memory Retrieval}
The memory retrieval system in Zep provides powerful, complex, and highly configurable functionality. At a high level, the Zep graph search API implements a function $f: S\to S$ that accepts a text-string query $\alpha\in S$ as input and returns a text-string context $\beta\in S$ as output. The output $\beta$ contains formatted data from nodes and edges required for an LLM agent to generate an accurate response to query $\alpha$. The process $f(\alpha)\to\beta$ comprises three distinct steps:

\begin{itemize}
\item \textbf{Search} ($\varphi$): The process begins by identifying candidate nodes and edges potentially containing relevant information. While Zep employs multiple distinct search methods, the overall search function can be represented as $\varphi: S\to\mathcal{E}_s^n\times\mathcal{N}_s^n\times\mathcal{N}_c^n$. Thus, $\varphi$ transforms a query into a 3-tuple containing lists of semantic edges, entity nodes, and community nodes—the three graph types containing relevant textual information.
\item \textbf{Reranker} ($\rho$): The second step reorders search results. A reranker function or model accepts a list of search results and produces a reordered version of those results: $\rho: {\varphi(\alpha),...}\to\mathcal{E}_s^n\times\mathcal{N}_s^n\times\mathcal{N}_c^n$.
\item \textbf{Constructor} ($\chi$): The final step, the constructor, transforms the relevant nodes and edges into text context: $\chi: \mathcal{E}_s^n\times\mathcal{N}_s^n\times\mathcal{N}c^n\to S$. For each $e_i\in\mathcal{E}s$, $\chi$ returns the fact and $t\text{valid},t\text{invalid}$ fields; for each $n_i\in\mathcal{N}_s$, the name and summary fields; and for each $n_i\in\mathcal{N}_c$, the summary field.
\end{itemize}
With these definitions established, we can express $f$ as a composition of these three components: $f(\alpha)=\chi(\rho(\varphi(\alpha)))=\beta$.

Sample context string template:

\noindent\fbox{%
    \parbox{\textwidth}{%
FACTS and ENTITIES represent relevant context to the current conversation.

These are the most relevant facts and their valid date ranges. If the fact is about an event, the event takes place during this time.

format: FACT (Date range: from - to)

<FACTS>

\{facts\}

</FACTS>

These are the most relevant entities

ENTITY\_NAME: entity summary

<ENTITIES>

\{entities\}

</ENTITIES>
    }%
}

\subsection{Search}
Zep implements three search functions: cosine semantic similarity search $(\varphi_\text{cos})$, Okapi BM25 full-text search $(\varphi_\text{bm25})$, and breadth-first search $(\varphi_\text{bfs})$. The first two functions utilize Neo4j's implementation of Lucene \cite{neo4j}\cite{lucene}. Each search function offers distinct capabilities in identifying relevant documents, and together they provide comprehensive coverage of candidate results before reranking. The search field varies across the three object types: for $\mathcal{E}_s$, we search the fact field; for $\mathcal{N}_s$, the entity name; and for $\mathcal{N}_c$, the community name, which comprises relevant keywords and phrases covered in the community. While developed independently, our community search approach parallels the high-level key search methodology in LightRAG \cite{lightrag}. The hybridization of LightRAG's approach with graph-based systems like Graphiti presents a promising direction for future research.

While cosine similarity and full-text search methodologies are well-established in RAG \cite{luceneisallyouneed}, breadth-first search over knowledge graphs has received limited attention in the RAG domain, with notable exceptions in graph-based RAG systems such as AriGraph \cite{arigraph} and Distill-SynthKG \cite{synthkg}. In Graphiti, the breadth-first search enhances initial search results by identifying additional nodes and edges within $n$-hops. Moreover, $\varphi_\text{bfs}$ can accept nodes as parameters for the search, enabling greater control over the search function. This functionality proves particularly valuable when using recent episodes as seeds for the breadth-first search, allowing the system to incorporate recently mentioned entities and relationships into the retrieved context.

The three search methods each target different aspects of similarity: full-text search identifies word similarities, cosine similarity captures semantic similarities, and breadth-first search reveals contextual similarities—where nodes and edges closer in the graph appear in more similar conversational contexts. This multi-faceted approach to candidate result identification maximizes the likelihood of discovering optimal context.

\subsection{Reranker}
While the initial search methods aim to achieve high recall, rerankers serve to increase precision by prioritizing the most relevant results. Zep supports existing reranking approaches such as Reciprocal Rank Fusion (RRF) \cite{rrf2009} and Maximal Marginal Relevance (MMR) \cite{mmr}. Additionally, Zep implements a graph-based episode-mentions reranker that prioritizes results based on the frequency of entity or fact mentions within a conversation, enabling a system where frequently referenced information becomes more readily accessible. The system also includes a node distance reranker that reorders results based on their graph distance from a designated centroid node, providing context localized to specific areas of the knowledge graph. The system's most sophisticated reranking capability employs cross-encoders—LLMs that generate relevance scores by evaluating nodes and edges against queries using cross-attention, though this approach incurs the highest computational cost.

\section{Experiments}
This section analyzes two experiments conducted using LLM-memory based benchmarks. The first evaluation employs the Deep Memory Retrieval (DMR) task developed in \cite{memgpt}, which uses a 500-conversation subset of the Multi-Session Chat dataset introduced in "Beyond Goldfish Memory: Long-Term Open-Domain Conversation" \cite{goldfishmemory}. The second evaluation utilizes the LongMemEval benchmark from "LongMemEval: Benchmarking Chat Assistants on Long-Term Interactive Memory" \cite{longmemeval}. Specifically, we use the LongMemEval$_{s}$ dataset, which provides an extensive conversation context of on average 115,000 tokens.

For both experiments, we integrate the conversation history into a Zep knowledge graph through Zep's APIs. We then retrieve the 20 most relevant edges (facts) and entity nodes (entity summaries) using the techniques described in Section 3. The system reformats this data into a context string, matching the functionality provided by Zep's memory APIs.

While these experiments demonstrate key retrieval capabilities of Graphiti, they represent a subset of the system's full search functionality. This focused scope enables clear comparison with existing benchmarks while reserving the exploration of additional knowledge graph capabilities for future work.

\subsection{Choice of models}
Our experimental implementation employs the BGE-m3 models from BAAI for both reranking and embedding tasks \cite{li2023making} \cite{bge-m3}. For graph construction and response generation, we utilize gpt-4o-mini-2024-07-18 for graph construction, and both gpt-4o-mini-2024-07-18 and gpt-4o-2024-11-20 for the chat agent generating responses to provided context.

To ensure direct comparability with MemGPT's DMR results, we also conducted the DMR evaluation using gpt-4-turbo-2024-04-09.

The experimental notebooks will be made publicly available through our GitHub repository, and relevant experimental prompts are included in the Appendix.

\subsection{Deep Memory Retrieval (DMR)}
The Deep Memory Retrieval evaluation, introduced by \cite{memgpt}, comprises 500 multi-session conversations, each containing 5 chat sessions with up to 12 messages per session. Each conversation includes a question/answer pair for memory evaluation. The MemGPT framework \cite{memgpt} currently leads performance metrics with 93.4\% accuracy using gpt-4-turbo, a significant improvement over the 35.3\% baseline achieved through recursive summarization.

To establish comparative baselines, we implemented two common LLM memory approaches: full-conversation context and session summaries. Using gpt-4-turbo, the full-conversation baseline achieved 94.4\% accuracy, slightly surpassing MemGPT's reported results, while the session summary baseline achieved 78.6\%. When using gpt-4o-mini, both approaches showed improved performance: 98.0\% for full-conversation and 88.0\% for session summaries. We were unable to reproduce MemGPT's results using gpt-4o-mini due to insufficient methodological details in their published work.

We then evaluated Zep's performance by ingesting the conversations and using its search functions to retrieve the top 10 most relevant nodes and edges. An LLM judge compared the agent's responses to the provided golden answers. Zep achieved 94.8\% accuracy with gpt-4-turbo and 98.2\% with gpt-4o-mini, showing marginal improvements over both MemGPT and the respective full-conversation baselines. However, these results must be contextualized: each conversation contains only 60 messages, easily fitting within current LLM context windows.

The limitations of the DMR evaluation extend beyond its small scale. Our analysis revealed significant weaknesses in the benchmark's design. The evaluation relies exclusively on single-turn, fact-retrieval questions that fail to assess complex memory understanding. Many questions contain ambiguous phrasing, referencing concepts like "favorite drink to relax with" or "weird hobby" that were not explicitly characterized as such in the conversations. Most critically, the dataset poorly represents real-world enterprise use cases for LLM agents. The high performance achieved by simple full-context approaches using modern LLMs further highlights the benchmark's inadequacy for evaluating memory systems.

This inadequacy is further emphasized by findings in \cite{longmemeval}, which demonstrate rapidly declining LLM performance on the LongMemEval benchmark as conversation length increases. The LongMemEval dataset \cite{longmemeval} addresses many of these shortcomings by presenting longer, more coherent conversations that better reflect enterprise scenarios, along with more diverse evaluation questions.

\begin{table}
\centering
\begin{threeparttable}
\caption{Deep Memory Retrieval}
\centering
\begin{tabular}{lll}
\toprule
Memory                        & Model         & Score      \\
\midrule
Recursive Summarization\tnote{†} & gpt-4-turbo  & 35.3\%     \\
Conversation Summaries         & gpt-4-turbo  & 78.6\%     \\
MemGPT\tnote{†}      & gpt-4-turbo  & 93.4\%     \\
Full-conversation              & gpt-4-turbo  & 94.4\%     \\
Zep                            & gpt-4-turbo  & \textbf{94.8\%} \\
\midrule
Conversation Summaries         & gpt-4o-mini  & 88.0\%     \\
Full-conversation              & gpt-4o-mini  & 98.0\%     \\
Zep                            & gpt-4o-mini  & \textbf{98.2\%} \\
\bottomrule
\end{tabular}
\begin{tablenotes}
\item[†] Results reported in \cite{memgpt}.
\end{tablenotes}
\end{threeparttable}
\label{tab:dmrtable}
\end{table}

\subsection{LongMemEval (LME)}
We evaluated Zep using the LongMemEval${s}$ dataset, which provides conversations and questions representative of real-world business applications of LLM agents. The LongMemEval${s}$ dataset presents significant challenges to existing LLMs and commercial memory solutions \cite{longmemeval}, with conversations averaging approximately 115,000 tokens in length. This length, while substantial, remains within the context windows of current frontier models, enabling us to establish meaningful baselines for evaluating Zep's performance.

The dataset incorporates six distinct question types: single-session-user, single-session-assistant, single-session-preference, multi-session, knowledge-update, and temporal-reasoning. These categories are not uniformly distributed throughout the dataset; for detailed distribution information, we refer readers to \cite{longmemeval}.

We conducted all experiments between December 2024 and January 2025. We performed testing using a consumer laptop from a residential location in Boston, MA, connecting to Zep's service hosted in AWS us-west-2. This distributed architecture introduced additional network latency when evaluating Zep's performance, though this latency was not present in our baseline evaluations. 

For answer evaluation, we employed GPT-4o with the question-specific prompts provided in \cite{longmemeval}, which have demonstrated high correlation with human evaluators.

\subsubsection{LongMemEval and MemGPT}
To establish a comparative benchmark between Zep and the current state-of-the-art MemGPT system \cite{memgpt}, we attempted to evaluate MemGPT using the LongMemEval dataset. Given that the current MemGPT framework does not support direct ingestion of existing message histories, we implemented a workaround by adding conversation messages to the archival history. However, we were unable to achieve successful question responses using this approach. We look forward to seeing evaluations of this benchmark by other research teams, as comparative performance data would benefit the broader development of LLM memory systems.

\subsubsection{LongMemEval results}

Zep demonstrates substantial improvements in both accuracy and latency compared to the baseline across both model variants. Using gpt-4o-mini, Zep achieved a 15.2\% accuracy improvement over the baseline, while gpt-4o showed an 18.5\% improvement. The reduced prompt size also led to significant latency cost reductions compared to the baseline implementations.

\begin{table}[h]
 \caption{LongMemEval$_{s}$}
  \centering
  \begin{tabular}{llllll}
    \toprule
    \multicolumn{2}{c}{}                  \\
    Memory     & Model     & Score  & Latency & Latency IQR  & Avg Context Tokens \\
    \midrule
    Full-context & gpt-4o-mini & 55.4\% & 31.3 s & 8.76 s & 115k\\
    Zep & gpt-4o-mini & \textbf{63.8\%} & \textbf{3.20 s} & 1.31 s & \textbf{1.6k} \\
    Full-context & gpt-4o & 60.2\% & 28.9 s & 6.01 s & 115k \\
    Zep & gpt-4o & \textbf{71.2\%} & \textbf{2.58 s} & 0.684 s & \textbf{1.6k}\\
    \bottomrule
  \end{tabular}
  \label{tab:lmeoverview}
\end{table}

Analysis by question type reveals that gpt-4o-mini with Zep showed improvements in four of the six categories, with the most substantial gains in complex question types: single-session-preference, multi-session, and temporal-reasoning. When using gpt-4o, Zep further demonstrated improved performance in the knowledge-update category, highlighting its effectiveness with more capable models. However, additional development may be needed to improve less capable models' understanding of Zep's temporal data.

\begin{table}[h]
 \caption{LongMemEval$_{s}$ Question Type Breakdown}
  \centering
  \begin{tabular}{lllll}
    \toprule
    \multicolumn{2}{c}{}                   \\
    Question Type & Model & Full-context & Zep  & Delta\\
    \midrule
    single-session-preference & gpt-4o-mini & 30.0\% & \textbf{53.3\%} & \textbf{77.7\%$\uparrow$}
    \\
    single-session-assistant & gpt-4o-mini & \textbf{81.8\%} & 75.0\% & 9.06\%$\downarrow$
    \\
    temporal-reasoning & gpt-4o-mini & 36.5\%  & \textbf{54.1\%} & \textbf{48.2\%$\uparrow$}
    \\
    multi-session & gpt-4o-mini & 40.6\% & \textbf{47.4\%} & \textbf{16.7\%$\uparrow$}
    \\
    knowledge-update & gpt-4o-mini & \textbf{76.9\%} & 74.4\% & 3.36\%$\downarrow$
    \\
    single-session-user & gpt-4o-mini & 81.4\%  & \textbf{92.9\%} & \textbf{14.1\%$\uparrow$}
    \\
    \cmidrule(r){1-5}
    single-session-preference & gpt-4o & 20.0\% & \textbf{56.7\%} & \textbf{184\%$\uparrow$}
    \\
    single-session-assistant & gpt-4o & \textbf{94.6\%} & 80.4\% & 17.7\%$\downarrow$
    \\
    temporal-reasoning & gpt-4o & 45.1\%  & \textbf{62.4\%} & \textbf{38.4\%$\uparrow$}
    \\
    multi-session & gpt-4o & 44.3\% & \textbf{57.9\%} & \textbf{30.7\%$\uparrow$}
    \\
    knowledge-update & gpt-4o & 78.2\% & \textbf{83.3\%} & \textbf{6.52\%$\uparrow$}
    \\
    single-session-user & gpt-4o & 81.4\%  & \textbf{92.9\%} & \textbf{14.1\%$\uparrow$}
    \\
    \bottomrule
  \end{tabular}
  \label{tab:lmefull}
\end{table}

These results demonstrate Zep's ability to enhance performance across model scales, with the most pronounced improvements observed in complex and nuanced question types when paired with more capable models. The latency improvements are particularly noteworthy, with Zep reducing response times by approximately 90\% while maintaining higher accuracy.

The decrease in performance for single-session-assistant questions—17.7\% for gpt-4o and 9.06\% for gpt-4o-mini—represents a notable exception to Zep's otherwise consistent improvements, and suggest further research and engineering work is needed.

\section{Conclusion}
We have introduced Zep, a graph-based approach to LLM memory that incorporates semantic and episodic memory alongside entity and community summaries. Our evaluations demonstrate that Zep achieves state-of-the-art performance on existing memory benchmarks while reducing token costs and operating at significantly lower latencies.

The results achieved with Graphiti and Zep, while impressive, likely represent only initial advances in graph-based memory systems. Multiple research avenues could build upon these frameworks, including integration of other GraphRAG approaches into the Zep paradigm and novel extensions of our work.

Research has already demonstrated the value of fine-tuned models for LLM-based entity and edge extraction within the GraphRAG paradigm, improving accuracy while reducing costs and latency \cite{synthkg}\cite{triplex}. Similar models fine-tuned for Graphiti prompts may enhance knowledge extraction, particularly for complex conversations. Additionally, while current research on LLM-generated knowledge graphs has primarily operated without formal ontologies \cite{arigraph}\cite{graphrag}\cite{lightrag}\cite{synthkg}\cite{graphreader}, domain-specific ontologies present significant potential. Graph ontologies, foundational in pre-LLM knowledge graph work, warrant further exploration within the Graphiti framework.

Our search for suitable memory benchmarks revealed limited options, with existing benchmarks often lacking robustness and complexity, frequently defaulting to simple needle-in-a-haystack fact-retrieval questions \cite{memgpt}. The field requires additional memory benchmarks, particularly those reflecting business applications like customer experience tasks, to effectively evaluate and differentiate memory approaches. Notably, no existing benchmarks adequately assess Zep's capability to process and synthesize conversation history with structured business data. While Zep focuses on LLM memory, its traditional RAG capabilities should be evaluated against established benchmarks such as those in \cite{lightrag}, \cite{financebench}, and \cite{beir}.

Current literature on LLM memory and RAG systems insufficiently addresses production system scalability in terms of cost and latency. We have included latency benchmarks for our retrieval mechanisms to begin addressing this gap, following the example set by LightRAG's authors in prioritizing these metrics.

\section{Appendix}
\subsection{Graph Construction Prompts}
\subsubsection{Entity Extraction}
\noindent\fbox{%
    \parbox{\textwidth}{%
<PREVIOUS MESSAGES>

\{previous\_messages\}

</PREVIOUS MESSAGES>

<CURRENT MESSAGE>

\{current\_message\}

</CURRENT MESSAGE>

Given the above conversation, extract entity nodes from the 
CURRENT MESSAGE that are explicitly or implicitly mentioned:

Guidelines:

1. ALWAYS extract the speaker/actor as the first node. The speaker 
is the part before the colon in each line of dialogue.

2. Extract other significant entities, concepts, or actors mentioned 
in the CURRENT MESSAGE.

3. DO NOT create nodes for relationships or actions.

4. DO NOT create nodes for temporal information like dates, times 
or years (these will be added to edges later).

5. Be as explicit as possible in your node names, using full names.

6. DO NOT extract entities mentioned only
    }%
}

\subsubsection{Entity Resolution}
\noindent\fbox{%
    \parbox{\textwidth}{%
<PREVIOUS MESSAGES>

\{previous\_messages\}

</PREVIOUS MESSAGES>

<CURRENT MESSAGE>

\{current\_message\}

</CURRENT MESSAGE>

<EXISTING NODES>

\{existing\_nodes\}

</EXISTING NODES>

Given the above EXISTING NODES, MESSAGE, and PREVIOUS MESSAGES. Determine if the NEW NODE extracted from the conversation
is a duplicate entity of one of the EXISTING NODES.

<NEW NODE>

\{new\_node\}

</NEW NODE>

Task:

1. If the New Node represents the same entity as any node in Existing Nodes, 
return 'is\_duplicate: true' in the 
    response. Otherwise, return 'is\_duplicate: false'

2. If is\_duplicate is true, also return the uuid of the existing node in the 
response

3. If is\_duplicate is true, return a name for the node that is the most complete 
full name.

Guidelines:

1. Use both the name and summary of nodes to determine if the entities are 
duplicates, duplicate nodes may have different names

    }%
}

\subsubsection{Fact Extraction}
\noindent\fbox{%
    \parbox{\textwidth}{%
<PREVIOUS MESSAGES>

\{previous\_messages\}

</PREVIOUS MESSAGES>

<CURRENT MESSAGE>

\{current\_message\}

</CURRENT MESSAGE>

<ENTITIES>

\{entities\}

</ENTITIES>

Given the above MESSAGES and ENTITIES, extract all facts pertaining to the 
listed ENTITIES from the CURRENT MESSAGE. 

Guidelines:

1. Extract facts only between the provided entities.

2. Each fact should represent a clear relationship between two DISTINCT nodes.

3. The relation\_type should be a concise, all-caps description of the 
fact (e.g., LOVES, IS\_FRIENDS\_WITH, WORKS\_FOR).

4. Provide a more detailed fact containing all relevant information.

5. Consider temporal aspects of relationships when relevant.

    }%
}

\subsubsection{Fact Resolution}
\noindent\fbox{%
    \parbox{\textwidth}{%
Given the following context, determine whether the New Edge represents any 
of the edges in the list of Existing Edges.

<EXISTING EDGES>

\{existing\_edges\}

</EXISTING EDGES>

<NEW EDGE>

\{new\_edge\}

</NEW EDGE>

Task:

1. If the New Edges represents the same factual information as any edge 
in Existing Edges, return 'is\_duplicate: true' in the response. 
Otherwise, return 'is\_duplicate: false'

2. If is\_duplicate is true, also return the uuid of the existing edge in 
the response

Guidelines:

1. The facts do not need to be completely identical to be duplicates, 
they just need to express the same information.

    }%
}

\subsubsection{Temporal Extraction}

\noindent\fbox{%
    \parbox{\textwidth}{%
<PREVIOUS MESSAGES>

\{previous\_messages\}

</PREVIOUS MESSAGES>

<CURRENT MESSAGE>

\{current\_message\}

</CURRENT MESSAGE>

<REFERENCE TIMESTAMP>

\{reference\_timestamp\}

</REFERENCE TIMESTAMP>
         
<FACT>

\{fact\}

</FACT>

IMPORTANT: Only extract time information if it is part of the provided fact. 
Otherwise ignore the time mentioned. 

Make sure to do your best to determine the dates if only the relative time is 
mentioned. (eg 10 years ago, 2 mins ago) based on the provided reference 
timestamp

If the relationship is not of spanning nature, but you are still able to 
determine the dates, set the valid\_at only.

         Definitions:

         - valid\_at: The date and time when the relationship described by the 
         edge fact became true or was established.

         - invalid\_at: The date and time when the relationship described by the 
         edge fact stopped being true or ended.

         Task:

         Analyze the conversation and determine if there are dates that are part 
         of the edge fact. Only set dates if they explicitly relate to the 
         formation or alteration of the relationship itself.

         Guidelines:

         1. Use ISO 8601 format (YYYY-MM-DDTHH:MM:SS.SSSSSSZ) for datetimes.

         2. Use the reference timestamp as the current time when determining 
         the valid\_at and invalid\_at dates.

         3. If the fact is written in the present tense, use the Reference 
         Timestamp for the valid\_at date

         4. If no temporal information is found that establishes or changes the 
         relationship, leave the fields as null.

         5. Do not infer dates from related events. Only use dates that are 
         directly stated to establish or change the relationship.

6. For relative time mentions directly related to the relationship, calculate the
actual datetime based on the reference timestamp.

         7. If only a date is mentioned without a specific time, use 00:00:00 (midnight) for that date.

         8. If only year is mentioned, use January 1st of that year at 00:00:00.

         9. Always include the time zone offset (use Z for UTC if no specific 
         time zone is mentioned).

    }%
}

%Bibliography
\bibliographystyle{unsrt}  
\bibliography{references}

\end{document}